\documentclass[runningheads]{llncs}
\usepackage{llncsdoc}
\usepackage{amsmath,amsfonts,amssymb}
\usepackage{bbm}
\usepackage{setspace}
\usepackage{graphicx}
\usepackage{color}
\usepackage{mathabx}
\usepackage{todonotes}
\usepackage{fancyhdr}

\usepackage{hyperref}
\hypersetup{
    pdfcreator={},
    pdfauthor={},
    pdfproducer={}
}

\usepackage{graphicx}

\newcommand{\by}{\boldsymbol{y}}
\newcommand{\bz}{\boldsymbol{z}}
\newcommand{\bh}{\boldsymbol{h}}
\newcommand{\bg}{g}
\newcommand{\bc}{c}
\newcommand{\bI}{I}
\newcommand{\bx}{\boldsymbol{x}}
\newcommand{\bu}{\boldsymbol{u}}

\newcommand{\bK}{\boldsymbol{K}}

\newcommand{\bW}{\boldsymbol{W}}

\newcommand{\bepsilon}{\boldsymbol{\epsilon}}

\newcommand{\balpha}{\boldsymbol{\alpha}}
\newcommand{\btheta}{\boldsymbol{\theta}}

\begin{document}

\title{Predictive Modeling of Anatomy \\with Genetic and Clinical Data}
\titlerunning{Predictive Modeling of Anatomy with Genetic and Clinical Data}
\authorrunning{A. Dalca et al.}

\author{Adrian V. Dalca\inst{1} \and Ramesh Sridharan\inst{1} \and Mert R. Sabuncu\inst{2} \and Polina Golland\inst{1} \\for the ADNI
	\thanks{A listing of ADNI investigators is available at http://tinyurl.com/ADNI-main. Data used via AD DREAM Challenge: https://www.synapse.org/\#!Synapse:syn2290704}}
\institute{Computer Science and Artificial Intelligence Lab, EECS, MIT \\
\and Martinos Center for Biomedical Imaging, Harvard Medical School}


\maketitle

\begin{abstract}
We present a semi-parametric generative model for predicting
anatomy of a patient in subsequent scans following a single 
baseline image.  Such predictive modeling promises to facilitate novel
analyses in both voxel-level studies and longitudinal biomarker
evaluation. We capture anatomical change through a combination of
population-wide regression and a non-parametric model of the
subject's health based on individual genetic and clinical indicators.
In contrast to classical correlation and longitudinal analysis, we focus on predicting new observations from a single subject observation. 
We demonstrate prediction of follow-up anatomical scans in the ADNI
cohort, and illustrate a novel analysis approach that compares a
patient's scans to the predicted subject-specific healthy anatomical trajectory.  The code is available at~\url{https://github.com/adalca/voxelorb}.


\end{abstract}


\thispagestyle{fancy}
\section{Introduction}

We present a method for predicting anatomy based on external information, including genetic and clinical indicators. Specifically, given only a single baseline scan of a new subject in a longitudinal study, our model predicts anatomical changes and generates a subsequent image by leveraging subject-specific genetic and clinical information. Such voxel-wise prediction opens up several new areas of analysis, enabling novel investigations both at the voxel level and at the level of derivative biomarker measures. For example, voxel level differences between the true progression of a patient with dementia and their predicted healthy anatomy highlight spatial patterns of disease. We validate our method by comparing measurements of volumes of anatomical structures based on predicted images to those extracted from the acquired scans.

Our model describes the change from a single (or \textit{baseline}) medical scan in terms of population trends and subject-specific external information. 
We model how anatomical appearance changes with age on average in a population, as well as deviations from the population average using a person's \textit{health profile}. We characterize such profiles non-parametrically based on the genotype, clinical information, and the baseline image. Subject-specific change is constructed from the similarity of health profiles in the cohort, using a Gaussian process parametrized by a population health covariance. Given the predicted change, we synthesize new images through an appearance model.

Statistical population analysis is one of the central topics in medical image computing. The classical correlation-based analysis has yielded important characterization of relationships within imaging data and with independent clinical variables~\cite{davis2010,misra2009,Pfefferbaum2013,rohlfing2009}. Regression models of object appearance have been previously used for atlas construction and population analysis~\cite{davis2010,rohlfing2009}. These methods characterize population trends with respect to external variables, such as age or gender, and construct clinically relevant population averages. 

Longitudinal analyses also characterize subject-specific temporal effects, usually in terms of changes in the biomarkers of interest. Longitudinal cohorts and studies promise to provide crucial insights into aging and disease~\cite{misra2009,Pfefferbaum2013}.
Mixed effects models have been shown to improve estimation of subject-specific longitudinal trends by using inter-population similarity~\cite{durrleman2013,sadeghi2013}. 
While these approaches offer a powerful basis for analysis of biomarkers or images in a population, they require multiple observations for any subject, and do not aim to provide subject-specific predictions given a single observation. The parameters of the models are examined for potential scientific insight, but they are not tested for predictive power.

Recently, several papers have proposed prediction of medical images or specific anatomical structures by extrapolating at least two time points (e.g. a baseline and a follow-up image) to simulate a later follow-up measurement~\cite{blanc2012,fleishman2015}. These models, however, necessitate observations of at least two time points to make predictions which are difficult to acquire for every new patient. They are therefore not applicable to most patients, not applicable during a first visit, and are sensitive to noise in these few measurements. Recent methods have also used stratified population average deformations to make follow-up predictions from a baseline image~\cite{modat2014}. While classified by population subgroups, these predictions are not adapted to a particular subject's health or environment. 

%
%
In contrast, we define the problem of population analysis as predicting anatomical changes for individual subjects. Our generative model incorporates a population trend and uses subject-specific genetic and clinical information, along with the baseline image, to generate subsequent anatomical images. This prediction-oriented approach provides avenues for novel analysis, as illustrated by our experimental results. 





\section{Prediction Model}

Given a dataset of patients with longitudinal data, and a single baseline image for a new patient, we predict follow-up anatomical states for the patient. We model anatomy as a phenotype~$y$ that captures the underlying structure of interest. For example,~$y$ can be a low-dimensional descriptor of the anatomy at each voxel. We assume we only have a measurement of our phenotype at baseline~$y_b$ for a new subject. Our goal is to predict the phenotype~$y_t$ at a later time~$t$. We let~$x_t$ be the subject age at time~$t$, and define~$\Delta x_t = x_t - x_b$ and~$\Delta y_t = y_t - y_b$. We model the change in phenotype~$y_t$ using linear regression:
\begin{align}
	\Delta y_t = \Delta x_t \beta + \epsilon,
	\label{eq:pred}
\end{align}
where~$\beta$ is the subject-specific regression coefficient, and noise~$\epsilon \sim \mathcal{N}(0, \sigma^2)$ is sampled from zero-mean Gaussian distribution with variance~$\sigma^2$.



\subsection{Subject-Specific Longitudinal Change}
To model subject-specific effects, we 
%
define~$\beta = \bar{\beta} + H(\bg,\bc,f_b)$, where~$\bar{\beta}$ is a global regression coefficient shared by the entire population, and~$H$ captures a deviation from this coefficient based on the subject's genetic variants~$g$, clinical information~$c$, and baseline image features~$f_b$.


\begin{figure}[t]
	\centering
	\begin{minipage}[t]{0.35\linewidth}
		\includegraphics[width=1\linewidth]{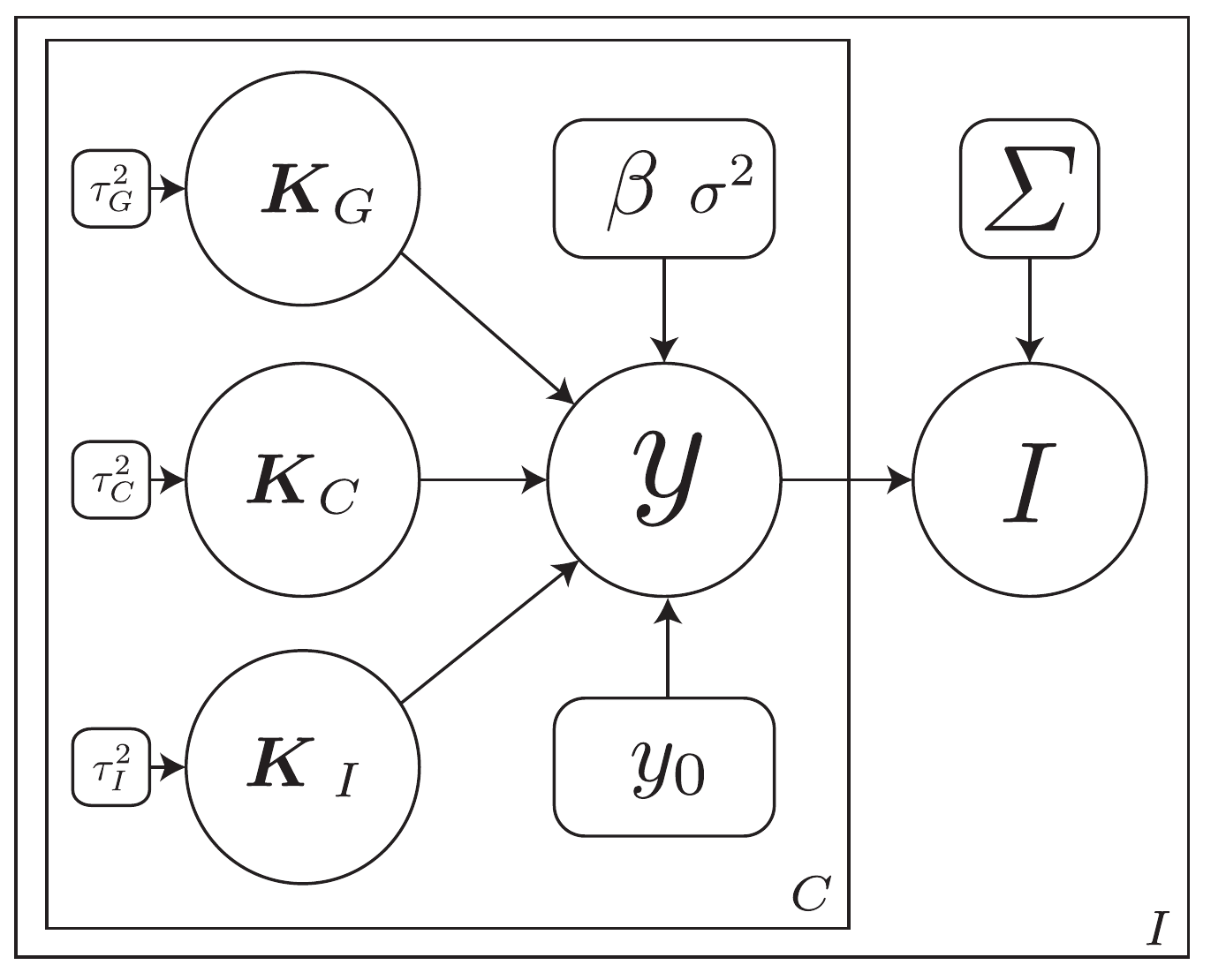}
	\end{minipage}
	\hfill
	\begin{minipage}[b]{0.60\linewidth}
		\caption{A graphical representation of our generative model. Circles indicate random variable and rounded squares represent parameters. The plates indicate replication.~$I$ is the predicted image, and~$y_c$ are the phenotypes that help us determine the image. Phenotypes are affected by~$h$, which are drawn from Gaussian processes and capture among-subject similarity through genetic, clinical and imaging factors.}
		\label{fig:graphicalmodel}
	\end{minipage}
\end{figure}

We assume that patients' genetic variants and clinical indicators affect their anatomical appearance, and that subjects with similar health profiles exhibit similar patterns of anatomical change. 
We let~$h_G(\cdot)$,~$h_C(\cdot)$,~$h_I(\cdot)$ be functions that capture genetic, clinical and imaging effects on the regression coefficients:
\begin{align}
	H(\bg, \bc, \bI_b) = h_G(\bg) + h_C(\bc) + h_I(f_b).
\end{align}
Combining with~\eqref{eq:pred}, we arrive at the full model
\begin{align}
	\Delta y_t = \Delta x_t \bar{\beta} + \Delta x_t \big( h_G(\bg) + h_C(\bc) + h_I(f_b) \big) + \epsilon,
	\label{eq:finalychange}
\end{align}
%
%
%
%
which captures the population trend~$\bar{\beta}$, as well as the subject-specific deviations
~$[h_G(\cdot), h_C(\cdot), h_I(\cdot)]$. 
Together with a description of an image from phenotype~$y$ in the next section, equations~\eqref{eq:finalychange} and~\eqref{eq:gp} define a generative probabilistic interpretation of our model, which we show graphically in Figure~\ref{fig:graphicalmodel}.

For a longitudinal cohort of~$N$ subjects, we group all~$T_i$ observations for subject~$i$ to form~$\Delta \by_{i} = [y_{i_1}, y_{i_2}, ... y_{i_{T_i}}]$. We then form the global vector~$\Delta \by = [\Delta \by_{1}, \Delta \by_{2}, ..., \Delta \by_{N}]$. We similarly form vectors~$\Delta \bx$,~$\bh_G$,~$\bh_C$,~$\bh_I$,~$\boldsymbol{g}$,~$\boldsymbol{c}$,~$\boldsymbol{f_b}$ and~$\bepsilon$, to build the full regression model:
\begin{align}
	\Delta \by &= \Delta \bx \bar{\beta} + \Delta \bx \odot \left(\boldsymbol{h}_G(\boldsymbol{g}) + \bh_C(\boldsymbol{c}) + \bh_I(\boldsymbol{f}_b) \right) + \bepsilon,
	\label{eq:population}
\end{align}
%
%
where~$\odot$ is the Hadamard, or element-wise product. This formulation is mathematically equivalent to a General Linear Model (GLM)~\cite{mccullagh1984} in terms of the health profile predictors~$[\boldsymbol{h}_G, \bh_C, \bh_I]$. 

We employ Gaussian process priors to model the health functions:
\begin{align}
	h_D(\cdot) \sim GP(\mathbf{0}, \tau_D^2 K_D(\cdot, \cdot)),
	\label{eq:gp}
\end{align}
where covariance kernel function~$\tau_D^2 K_D(z_i, z_j)$ captures the similarity between subjects~$i$ and~$j$ using feature vectors~$\bz_i$ and~$\bz_j$ for~$D\in\{G,C,I\}$. We discuss the particular form of~$K(\cdot, \cdot)$ used in the experiments later in the paper.

\subsection{Learning}

The Bayesian formulation in~\eqref{eq:population} and~\eqref{eq:gp} can be interpreted as a linear mixed effects model (LMM)~\cite{mcculloch2001} or a least squares kernel machine (LSKM) regression model~\cite{ge2015,liu2007}. 
We use the LMM interpretation to learn the parameters of our model, and the LSKM interpretation to perform final phenotype predictions. 

Specifically, we treat~$\bar{\beta}$ as the coefficient vector of fixed effects and~$\bh_G, \bh_C$, and~$\bh_I$ as independent random effects. We seek the maximum likelihood estimates of parameters~$\bar{\beta}$ and~$\btheta = (\tau^2_G, \tau^2_C, \tau^2_I, \sigma^2)$ by adapting standard procedures for LMMs~\cite{ge2015,liu2007}. As standard LMM solutions become computationally expensive for thousands of observations, we take advantage of the fact that while the entire genetic and the image phenotype data is large, the use of kernels on baseline data reduces the model size substantially. We obtain intuitive iterative updates 
%
%
that project the residuals at each step onto the expected rate of change in likelihood, and
%
update~$\bar{\beta}$ using the best linear unbiased predictor.
%
%

\subsection{Prediction}

Under the LSKM interpretation, the terms~$h(\cdot)$ are estimated by minimizing a penalized squared-error loss function, which leads to the following solution~\cite{ge2015,kimeldorf1971,liu2007,wahba1990}:
\begin{align}
	h(z_i) = \sum_{j=1}^N \alpha_j K(\bz_i,\bz_j) \quad \mathrm{or} \quad \bh = \balpha^T \bK
\end{align}
for some vector~$\balpha$. Combining with the definitions of the LMM, we estimate coefficients vectors~$\balpha_G,\balpha_C$ and~$\balpha_I$ 
from a linear system of equations that involves our estimates of~$\hat{\beta}$ and~$\mathbf{\theta}$.
%
We can then re-write~\eqref{eq:population} as
\begin{align}
	\Delta \by = \Delta \bx \bar{\beta} + \Delta \bx \left(\balpha_G^T \bK_G + \balpha_C^T \bK_C + \balpha_I^T \bK_I  \right)
\end{align}
and predict a phenotype at time~$t$ for a new subject~$i$:
\begin{align}
	y_t = y_b + \Delta x_t \left[\bar{\beta} + \sum_{j=1}^N \alpha_{G,j} K_G(\bg_i,\bg_j) + \alpha_{C,j} K_C(\bc_i,\bc_j) + \alpha_{I,j} K_I(f_i,f_j)  \right].
	\label{eq:finalypred}
\end{align}

%


\section{Model Instantiation for Anatomical Predictions}
The full model~\eqref{eq:finalychange} can be used with many reasonable phenotype definitions. Here, we describe the phenotype model we use for anatomical predictions and specify the similarity kernels of the health profile.

\subsection{Anatomical Phenotype}
\label{sec:appearancemodel}

We define a voxel-wise phenotype that enables us to predict entire anatomical images.  Let~$\Omega$ be the set of all spatial locations~$v$ (voxels) in an image, and~$I_b = \{I_b(v)\}_{v\in\Omega}$ be the acquired baseline image. We similarly define~$A = \{A(v)\}_{v\in\Omega}$, to be the population atlas template. We assume each image~$I$ is generated through a deformation field~$\Phi_{AI}^{-1}$ parametrized by the corresponding displacements~$\{\bu(v)\}_{v\in\Omega}$ from the common atlas to the subject-specific coordinate frame~\cite{rohlfing2009}, such that~$
	I(v) = A(v + \bu(v)).
$
We further define a follow-up image~$I_t$ as a deformation~$\Phi_{Bt}$ from the baseline image~$I_b$, which can be composed to yield an overall deformation from the atlas to the follow-up scan via~$\Phi^{-1}_{At}=\Phi^{-1}_{AB} \circ \Phi^{-1}_{Bt} = \{\bu'(v)\}_{v\in\Omega}$:
\begin{align}
%
	I_t(v) = A(v + \bu'(v)).
\end{align}

\begin{figure}[t]
	\centering
	\includegraphics[width=1.0\linewidth]{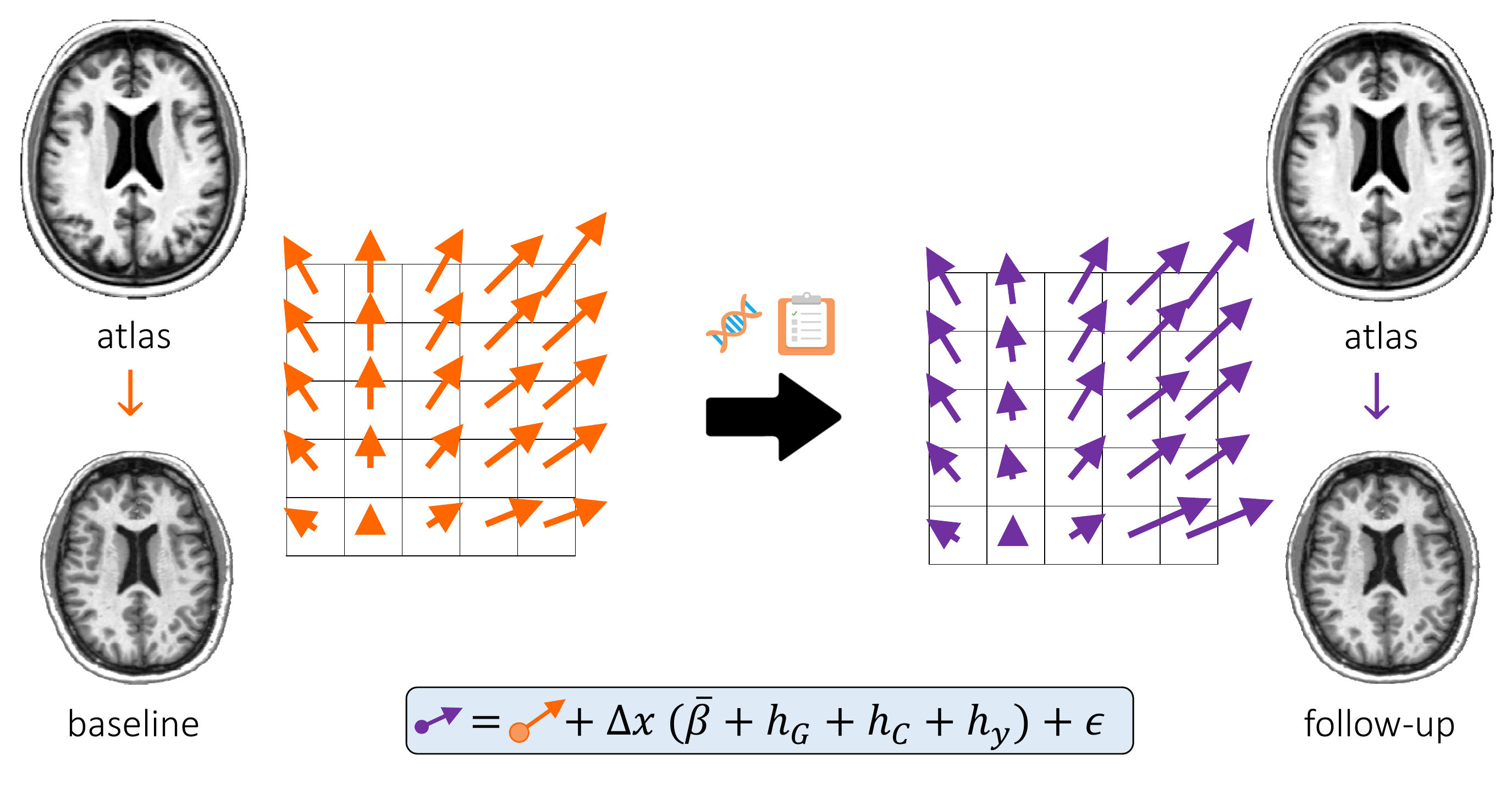}
	\caption{Overview of anatomical prediction. Arrows represent a displacement vector at each voxel from the common atlas space to the subject space. We predict the change in a displacement field from the baseline (orange) to the follow-up timepoint (purple) using external factors. A follow-up image can then be formed by propagating voxels from the baseline, to the atlas, and finally to the follow-up image locations.}
	\label{fig:schematic_anatomicalprediction}
\end{figure}

Using displacements~$\bu'(v)$ as the phenotype of interest in~\eqref{eq:pred} captures the necessary information for predicting new images, but leads to very high dimensional descriptors. To regularize the transformation and to improve efficiency, we define a low-dimensional embedding of~$\bu'(v)$. Specifically, we assume that the atlas provides a parcellation of the space into L anatomical labels~$\mathcal{L} = \{\Psi\}_{l=1}^L$. We build a low-dimensional embedding of the transformation vectors~$\bu(v)$ within each label using PCA. We define the relevant phenotypes~$\{y_{l,c}\}$ as the coefficients associated with the first~$C$ principal components of the model that capture 95\% of the variance in each label, for~$l=1\ldots L$.

We predict the phenotypes using~\eqref{eq:finalypred}. To construct a follow-up image~$I_t$ given phenotype~$y_t$, we first form a deformation field~$\widehat{\Phi}_{At}^{-1}$ by reconstruction from the estimated phenotype~$y_t$, and use~$\widehat{\Phi}_{At}$ assuming an invertible transformation. Using the baseline image, we predict a subsequent image via~$\Phi_{Bt} = \widehat{\Phi}_{At}\circ\Phi^{-1}_{AB}$. Note that we do not directly model changes in image intensity. While population models necessitate capturing such changes, we predict changes from a baseline image. We also assume that affine transformations are not part of the deformations of interest, and thus all images are affinely registered to the atlas. 

Using this appearance model, we use equation~\eqref{eq:finalypred} to predict the anatomical appearance at every voxel, as illustrated in Figure~\ref{fig:schematic_anatomicalprediction}.

\subsection{Health Similarities}
To fully define the health similarity term~$H(\cdot, \cdot, \cdot)$, we need to specify the forms of the kernel functions~$K_G(\cdot, \cdot)$,~$K_C(\cdot, \cdot)$, and~$K_I(\cdot, \cdot)$.

For genetic data, we employ the identical by state (IBS) kernel often used in genetic analysis~\cite{queller1989}. Given a vector of genetic variants~$\bg$ of length~$S$, each genetic locus is encoded as~$\bg(s) \in \{0, 1, 2\}$, and
%

\begin{equation}
	K_G(\bg_i, \bg_j) = \frac{1}{2S} \sum_{s=1}^S (2-|\bg_i(s) - \bg_{j}(s)|).
\end{equation}

To capture similarity of clinical indicators~$\bc$, we  
form the kernel function

\begin{equation}
	K_C(\bc_i, \bc_j) = \exp \left( -\frac{1}{\sigma^2_C}  (\bc_i - \bc_j)^T\bW(\bc_i - \bc_j) \right),
\end{equation}
where diagonal weight matrix~$\bW$ captures the effect size of each clinical indicator on the phenotype, and~$\sigma^2_C$ is the variance of the clinical factors. 

We define the image feature vectors $f_b$ as the set of all PCA coefficients defined above for the baseline image. We define the image kernel matrix as 
\begin{equation}
	K_I(f_{b,i}, f_{b,j}) = \exp \left( -\frac{1}{\sigma^2_I}  ||f_{b,i} - f_{b,j}||^2_2 \right),
\end{equation}
where~$\sigma^2_I$ is the variance of the image features.

We emphasize that other data sources can easily be incorporated in the model by simply adding appropriate kernels for that data.

\section{Experiments}
\begin{figure}[t]
\centering
\includegraphics[width=0.97\linewidth]{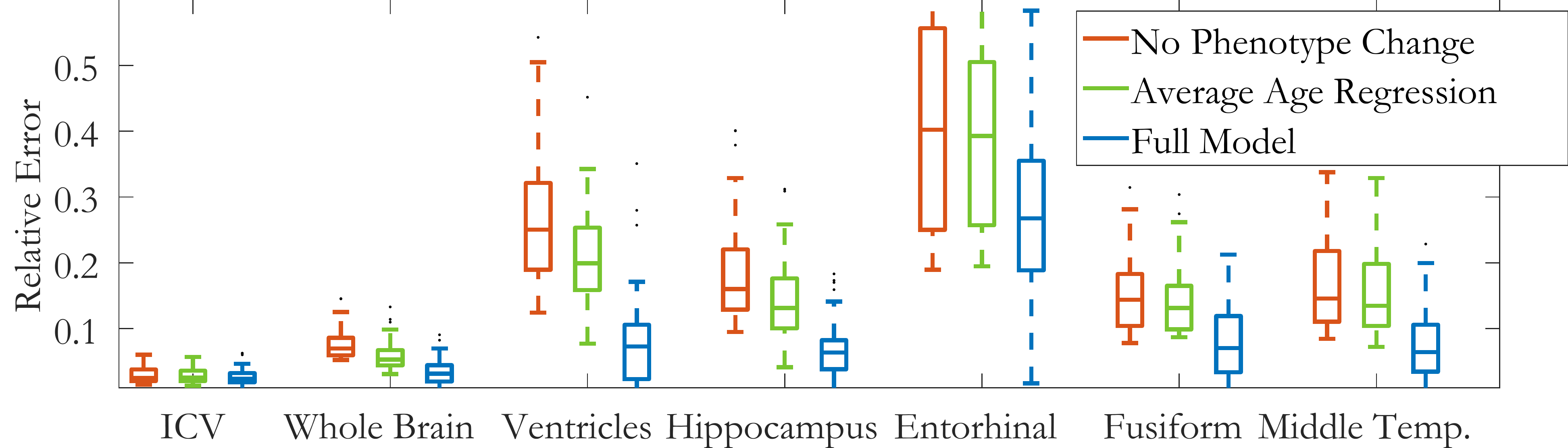}
\caption{\footnotesize Relative error (lower is better) of volume prediction for seven structures for subjects in the top decile of volume change. 
We report relative change between the baseline and the follow-up measurement (red), relative error in prediction using a population model (green), and the complete model (blue).}
\label{fig:volumeerror}
\end{figure}

\begin{figure}[b]
	\centering
	\includegraphics[width=0.97\linewidth]{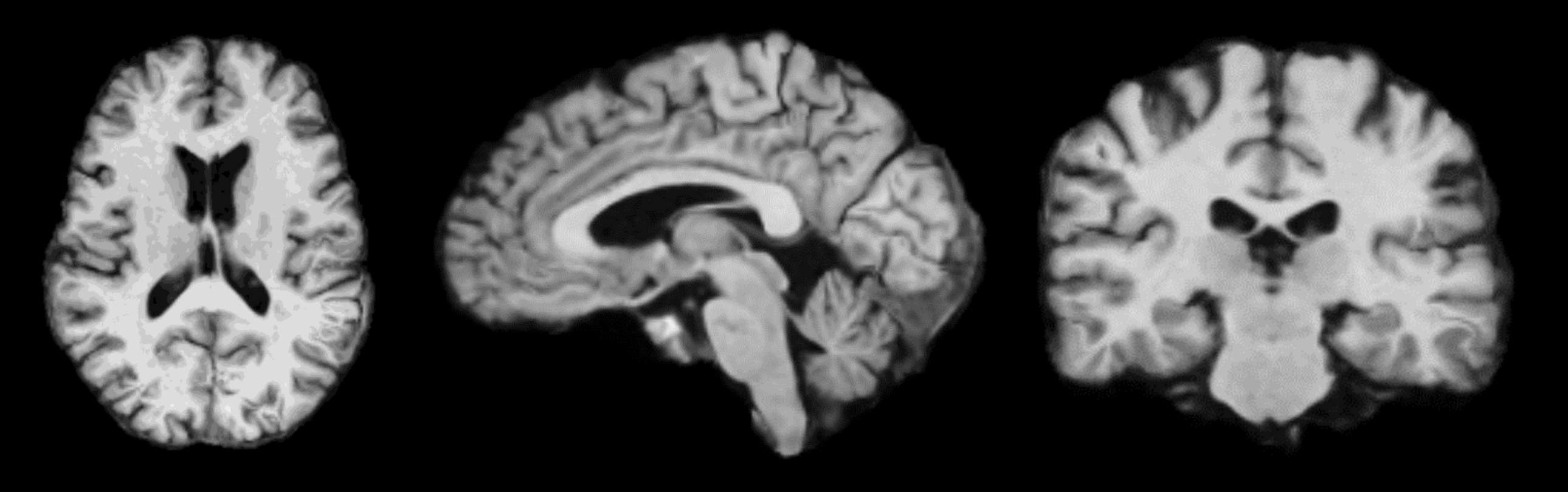}
	\caption{Example predicted volume, in axial, sagittal and coronal views. Qualitatively, the scan presents plausible anatomy, and would be challenging to discern from an actual scan.}
	\label{fig:examplesynthesis}
\end{figure}

\begin{figure}[tb]
	\centering
	\includegraphics[width=0.97\linewidth]{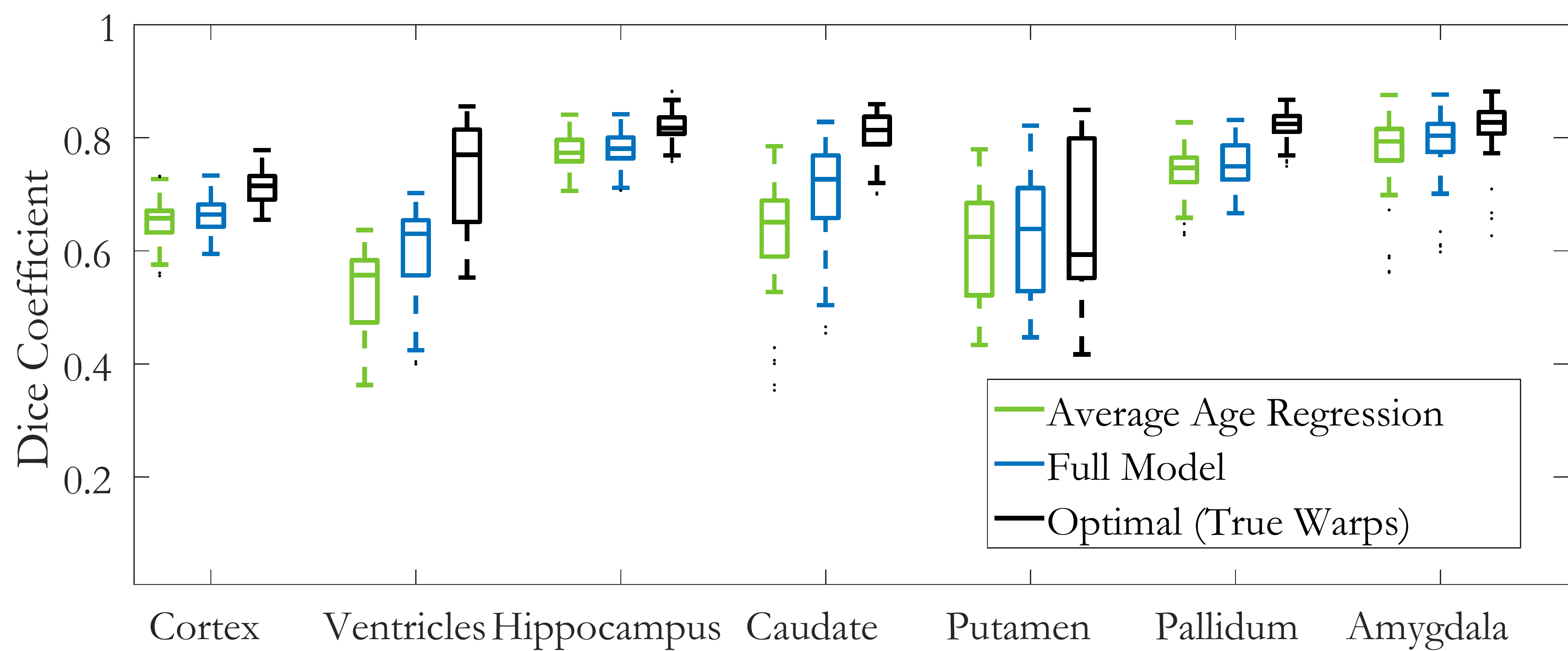}
	\caption{Prediction results. Dice scores of labels propagated through three methods for several structures implicatd in AD in subjects with the most volume change for each structure. We report the prediction based on the registration of the actual follow-up scan to the atlas as an upper bound for warp-based prediction accuracy (black), predictions based on the population-wide linear regression model (green), and the full model (blue).} 
	\label{fig:ADDice}
\end{figure}

We illustrate our approach by predicting image-based phenotypes based
on genetic, clinical and imaging data in the ADNI longitudinal
study~\cite{jack2008}
that includes two to
ten follow-up scans acquired~$0.5 - 7$ years after the baseline scan.  We use affine registration to align all subjects to
a template constructed from~$145$ randomly chosen subjects, and
compute non-linear registration warps~$\Phi_{AI}$ for each image using
ANTs~\cite{avants2011}. We utilize a list of~$21$ genetic loci
associated with Alzheimer's disease (AD) as the genetic vector $g$, and the 
standard clinical factors including age, gender, marital status,
education, disease diagnostic, and cognitive tests, as the clinical
indicator vector $c$. We learn the model parameters from 341 randomly chosen subjects
and predict follow-up volumes on a separate set of 100 subjects. To
evaluate the advantages of the proposed predictive model, we compare
its performance to a population-wide linear regression model that
ignores the subject-specific health profiles (i.e.,~$H=0$).

\subsection{Volumetric Predictions}

In the first simplified experiment, we define phenotype $y$ to be a
vector of several scalar volume measurements obtained using
FreeSurfer~\cite{fischl2012}. 
In addition to the population-wide
linear regression model, we include a simple approach of using the
baseline volume measurements as a predictor of the phenotype trajectory, effectively assuming no volume change with
time. Since in many subjects, the volume differences are
small, all three methods perform comparably when
evaluated on the whole test set. To evaluate the differences between
the methods, we focus on the subset of subjects with substantial
volume changes, reported in Fig.~\ref{fig:volumeerror}. 
Our method consistently achieves smaller relative errors than the
two baseline approaches.


\subsection{Anatomical Prediction}

\begin{figure}[t]
	\centering
	\includegraphics[width=0.97\linewidth]{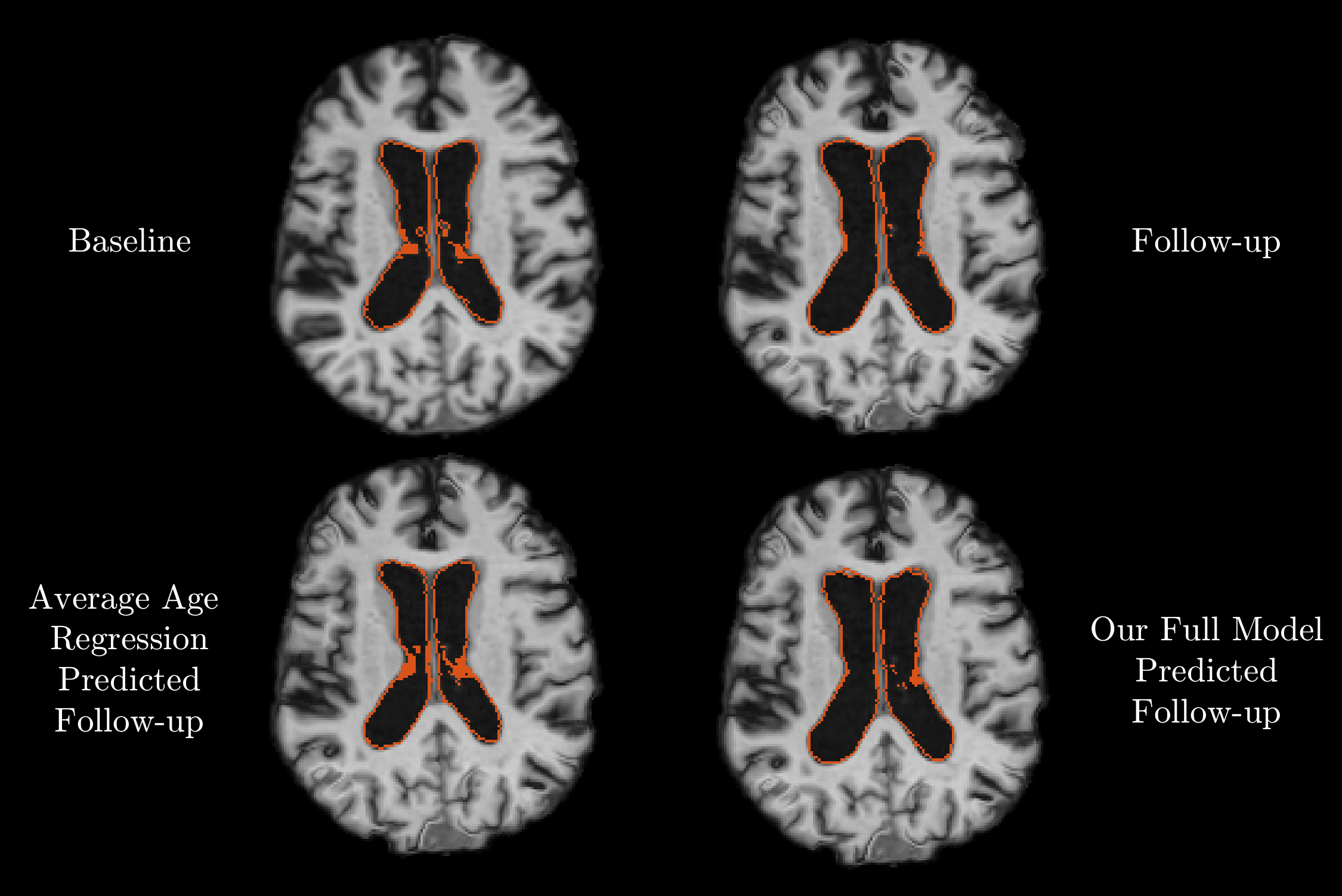}
	\caption{Example comparison of predicted follow-up scans. This subjects presents with significant ventricle expansion due to brain atrophy from the baseline image to the follow-up observation. The predicted scan using the full model is able to capture significantly more of the ventricle expansion compared to the age-regression based follow-up. Qualitatively, both predicted follow-up scans present as plausible anatomical images. 
	}
	\label{fig:comparisoneg}
\end{figure}

We also evaluate the model for full anatomical scan prediction. An example prediction is shown in Fig.~\ref{fig:examplesynthesis}. To quantify prediction accuracy, we propagate segmentation labels of
relevant anatomical structures from the baseline scan to the predicted
scan using the predicted warps. We compare the predicted
segmentation label maps with the actual segmentations of the follow-up
scans. The warps computed based on the actual follow-up
scans through the atlas provide an indication of the best accuracy the predictive model
could achieve when using warps to represent images. Similar to the
volumetric predictions, the full model offers modest improvements when
evaluated on the entire test set, and substantial improvements in 
segmentation accuracy when evaluated in the subjects who exhibit large
volume changes between the baseline scan and the follow-up scan, as
reported in Fig.~\ref{fig:ADDice} and exemplified in Fig.~\ref{fig:comparisoneg}.
In both experiments, all components~$h_g, h_c$ and~$h_I$ contributed significantly to the improved predictions.

\begin{figure}
	\centering
	\includegraphics[width=0.9\linewidth]{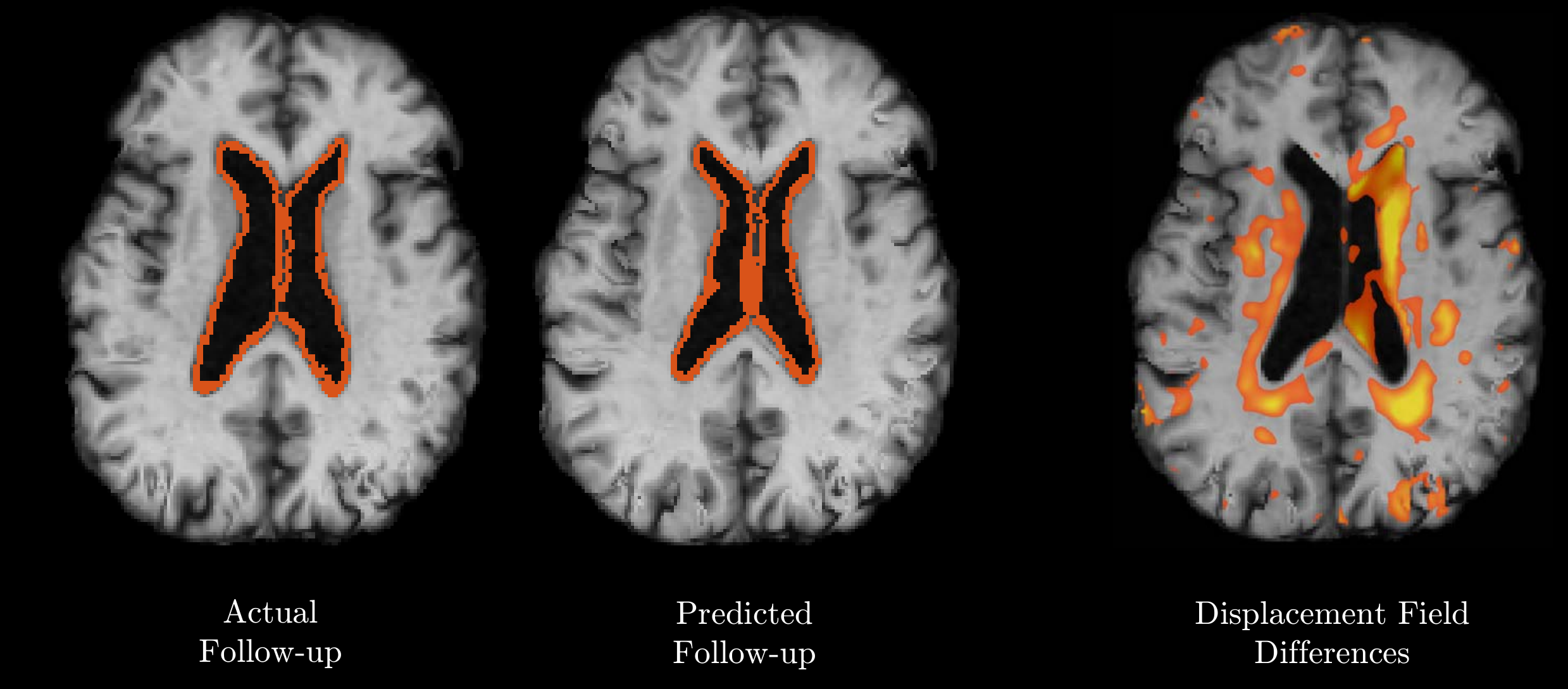}
	\includegraphics[width=0.052\linewidth]{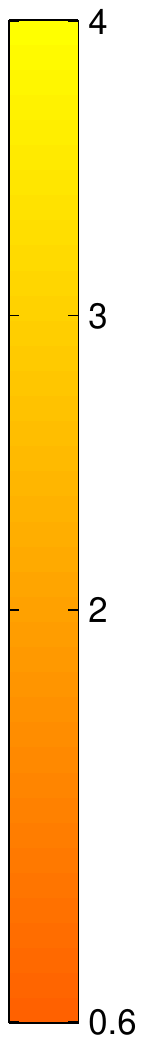}
	\caption{Comparison of an acquired and predicted anatomical image for a patient diagnosed with AD using a healthy model. Ventricles are outlines in red. The third image shows the predicted follow-up with a color overlay, indicating the squared magnitude of the difference in predicted versus observed deformation field, showing that the ventricles expand significantly less in the predicted scan. The figures indicate a significantly different expansion trajectory of the ventricles when compared to the observed follow-up.} 
	\label{fig:exampleADNI}
\end{figure}

Our experimental results suggest that the anatomical model depends on
registration accuracy.  In particular, we observe that directly
registering the follow-up scan to the baseline scan leads to better
alignment of segmentation labels than when transferring the labels
through a composition of the transformations from the scans to the
atlas space. This suggests that a different choice of appearance model
may improve prediction accuracy, a promising direction for future
work.

\subsection{Example Application}

To demonstrate the potential of the anatomical prediction, we predict
the follow-up scan of a patient diagnosed with dementia as if the
patient were healthy. Specifically, we train our model using healthy
subjects, and predict follow-up scans for AD patients.  In
Fig.~\ref{fig:exampleADNI} we illustrate an example result, comparing
the areas of brain anatomy that differ from the observed follow-up in the predicted
\textit{healthy} brain of this AD patient. Our prediction indicates that
ventricle expansion would be different if this patient had a healthy
trajectory.





\section{Conclusions}
%
%

We present a model to predict the anatomy in patient follow-up
images given just a baseline image using population trends and subject-specific genetic and
clinical information. We validate our prediction method on scalar volumes
and anatomical images, and show that it can be used as a powerful
tool to illustrate how a subject-specific brain might differ if it
were healthy. Through this and other new applications, our prediction
method presents a novel opportunity for the study of disease and
anatomical development.

\vspace{0.1cm}
\noindent {\footnotesize \textbf{Acknowledgements.} We acknowledge the following funding sources: NIH NIBIB 1K25EB013649-01, BrightFocus AHAF-A2012333, NIH NIBIB NAC P41EB015902, NIH DA022759, and Wistron Corporation.}

\scriptsize
\bibliography{refs}
\bibliographystyle{splncs03}


\end{document}